\documentclass[letterpaper, 10 pt, conference]{ieeeconf}  % Comment this line out if you need a4paper

\IEEEoverridecommandlockouts                              % This command is only needed if 
\usepackage{amsmath} % assumes amsmath package installed
\usepackage{graphicx}
\usepackage{pgfplots}
\usepackage{subcaption}
\usepackage{float}
\usepackage{tabularx}

\title{\LARGE \bf
F1tenth Autonomous Racing With Offline Reinforcement Learning Methods
}

\author{Prajwal Koirala$^{1}$ and Cody Fleming$^{2}$% <-this % stops a space
\thanks{$^{1, 2}$ The authors are affiliated with Iowa State University, Ames, Iowa.}%
}

\begin{document}

\maketitle
\thispagestyle{empty}
\pagestyle{empty}

%%%%%%%%%%%%%%%%%%%%%%%%%%%%%%%%%%%%%%%%%%%%%%%%%%%%%%%%%%%%%%%%%%%%%%%%%%%%%%%%
\begin{abstract}
Autonomous racing serves as a critical platform for evaluating automated driving systems and enhancing vehicle mobility intelligence. This work investigates offline reinforcement learning methods to train agents within the dynamic F1tenth racing environment. The study begins by exploring the challenges of online training in the Austria race track environment, where agents consistently fail to complete the laps. Consequently, this research pivots towards an offline strategy, leveraging `expert' demonstration dataset to facilitate agent training. A waypoint-based suboptimal controller is developed to gather data with successful lap episodes. This data is then employed to train offline learning-based algorithms, with a subsequent analysis of the agents' cross-track performance, evaluating their zero-shot transferability from seen to unseen scenarios and their capacity to adapt to changes in environment dynamics. Beyond mere algorithm benchmarking in autonomous racing scenarios, this study also introduces and describes the machinery of our return-conditioned decision tree-based policy, comparing its performance with methods that employ fully connected neural networks, Transformers, and Diffusion Policies and highlighting some insights into method selection for training autonomous agents in driving interactions.
\end{abstract}

%%%%%%%%%%%%%%%%%%%%%%%%%%%%%%%%%%%%%%%%%%%%%%%%%%%%%%%%%%%%%%%%%%%%%%%%%%%%%%%%
\section{INTRODUCTION}

%\begin{itemize}
%    \item We apply the RTG conditioned offline learning with XGB to F1tenth. Why?
%    \item We train and then test cross-track generalization. Why?
%    \item We compare to DT
%    item Compare to online RL
%\end{itemize}

%\begin{itemize}
%    \item 
%    \item F1tenth is a good platform for testing and developing autonomous systems; cheaper, safer, etc...and we also have sim capability
%    \item The latter is advantageous for testing RL algorithms (real-world testing is challenging)
%\end{itemize}
Methods for decision-making in autonomous racing vehicles can be broadly categorized into two main types: rule-based and learning-based approaches. Rule-based methods, while offering a structured framework, possess inherent limitations due to their reliance on manually crafted models, often imposing unrealistic assumptions. Their development needs substantial efforts in engineering and validation, but they still tend to exhibit poor adaptability to unseen scenarios. This motivates the exploration of learned models as potential alternatives for autonomous racing, given the intricate nature of driving interactions and the necessity for generalization \cite{huval2015empiricaldeeplearning, leon2019reviewautonomousdriving, zhang2024surveyracing}. These methods utilize computational techniques (like machine learning) to learn how to extract patterns, insights and intricacies from data and `autonomously’ improve their performance over time/experience.

Among learning-based methods, model-free approaches adopt a data-centric strategy, aiming to learn predictive distributions over the action set directly from available data. In contrast, the model-based approaches either learn a predictive distribution over state transitions or often leverage prior knowledge (like vehicle kinematics, maneuvers, scenarios, etc) to forecast realistic maneuvers and trajectories. By incorporating pre-existing information, model-based methods aim to enhance the efficiency of predictions, especially in situations where limited data is available. Although the model-free methods show greater flexibility by reducing the necessity of extensive prior knowledge, the large amount of data required might prove prohibitive. Thus, the current model-free learning-based methodologies, despite their potential, have not yet attained the level of safety and adaptability for autonomous racing/driving \cite{bosello2022traininaustria, djuric2020uncertaintyawareplanning, leon2019reviewautonomousdriving, betz2022autonomous, yalamanchi2020long}.

Learning to control a racing vehicle through online methods, however, poses significant challenges, including the risk of potential accidents during the agent's learning process and the logistical demands of continuous interaction with the environment, particularly in real-world applications like autonomous racing \cite{lange2012batch, fujimoto2021minimalist, prudencio2023survey}. Additionally, our findings, as explained in the subsequent sections, along with the research conducted by \cite{brunnbauer2021model}, underscore the limitations of model-free online learning, often faltering to learn the intricate interactions when navigating demanding racing circuits. 
In contrast, the offline reinforcement learning (Offline RL) paradigm, in which the agents are trained with pre-existing datasets or expert demonstrations, bypasses direct interaction with the environment, thus mitigating risks and failures associated with learning from scratch \cite{levine2020offline, tarasov2024corl}. With the prospect of leveraging extensive datasets of human driving behaviors to train agents, the offline learning of driving policies stands as a pivotal research direction within intelligent transportation. 

% In contrast, offline learning of policies addresses these challenges by mitigating these risks and failures that arise due to learning from scratch. Offline reinforcement learning facilitates training agents using a pre-existing dataset or expert demonstrations, thus bypassing the necessity for direct interaction with the environment and therefore reducing the risks associated with untrained agents.

The F1tenth platform serves as a compelling arena for the examination and validation of reinforcement learning algorithms in autonomous racing. The F1tenth platform has an associated virtual environment that provides a cost-effective, expedited, and secure method for deploying and assessing these algorithms. Specifically, this simulator offers a means to scrutinize the efficacy of autonomous racing algorithms before their implementation on the physical F1/10 racecar \cite{o2020f1tenthrl}. Advantageous for reinforcement learning, the virtual environment mitigates the expense of numerous failed episodes involving potential racecar crashes before achieving a proficient policy capable of safely completing laps. This paper specifically makes use of the Gym API for F1tenth racing simulator \cite{Axel2021racecar, brunnbauer2021model} developed using PyBullet physics simulation engine \cite{coumans2016pybullet}.

In our approach to F1tenth racing using offline RL methods, we begin by collecting data with a classical controller tailored explicitly for a racing track/map. This controller does not rely on lidar data but rather utilizes map, odometry, and waypoints to guide the racecar through the racetrack. Then, we train our offline RL policies using those `expert' demonstrations to predict actions--motor thrust and steering rate--based on observation space that only consists of lidar and velocity information. Subsequent to training, the cross-track generalization of each policy is evaluated by testing its performance on other maps.

Our previous work proposed Return-Conditioned Decision Tree Policy (RCDTP), a deterministic policy structured as an ensemble of weak learners that are conditioned on input states, return-to-go (RTG) and timestep information \cite{koirala2024reframing}.
The use of `extreme' gradient-boosted trees \cite{chen2016xgboost} in this model-free approach facilitated swift agent training and inference, enabling rapid experimentation and real-time decision-making capabilities. 
The primary contribution of the present study is to offer an exposition of this methodology within the context of the F1tenth racing domain and conduct a rigorous evaluation of its performance against other offline RL methods.
We have employed RL methods that feature large-scale state-of-the-art architectures such as Decision Transformer \cite{chen2021decision} and Diffusion Policy \cite{chi2023diffusion}, which utilize Transformer \cite{vaswani2017attention} and DDPM (Denoising Diffusion Probabilistic Models) \cite{ho2020denoising} respectively. Other popular offline RL baselines that use various variants of Q-learning like IQL \cite{kostrikov2021offline}, CQL \cite{kumar2020conservative}, AWAC \cite{nair2020awac}, PLAS (with perturbation) \cite{zhou2021plas}, and TD3+BC \cite{fujimoto2021minimalist} that use fully-connected neural network (FCNN) have also been implemented below. These methods are implemented in the F1tenth racing context using the d3rlpy library, which provides API for offline RL algorithms \cite{seno2022d3rlpy}. Among the offline methods, our comparative analysis underscores better sample efficiency and rapid training speed of RCDTP, robustness of generative modeling-based methods and better generalization of large-architecture models in cross-track datasets.
Additionally, we also assess the state-of-the-art online reinforcement learning algorithms such as SAC \cite{haarnoja2018softactor} and PPO \cite{schulman2017proximal} for comprehensive assessment and benchmarking purposes.

\begin{figure}
    \centering
    \includegraphics[width=0.9\linewidth]{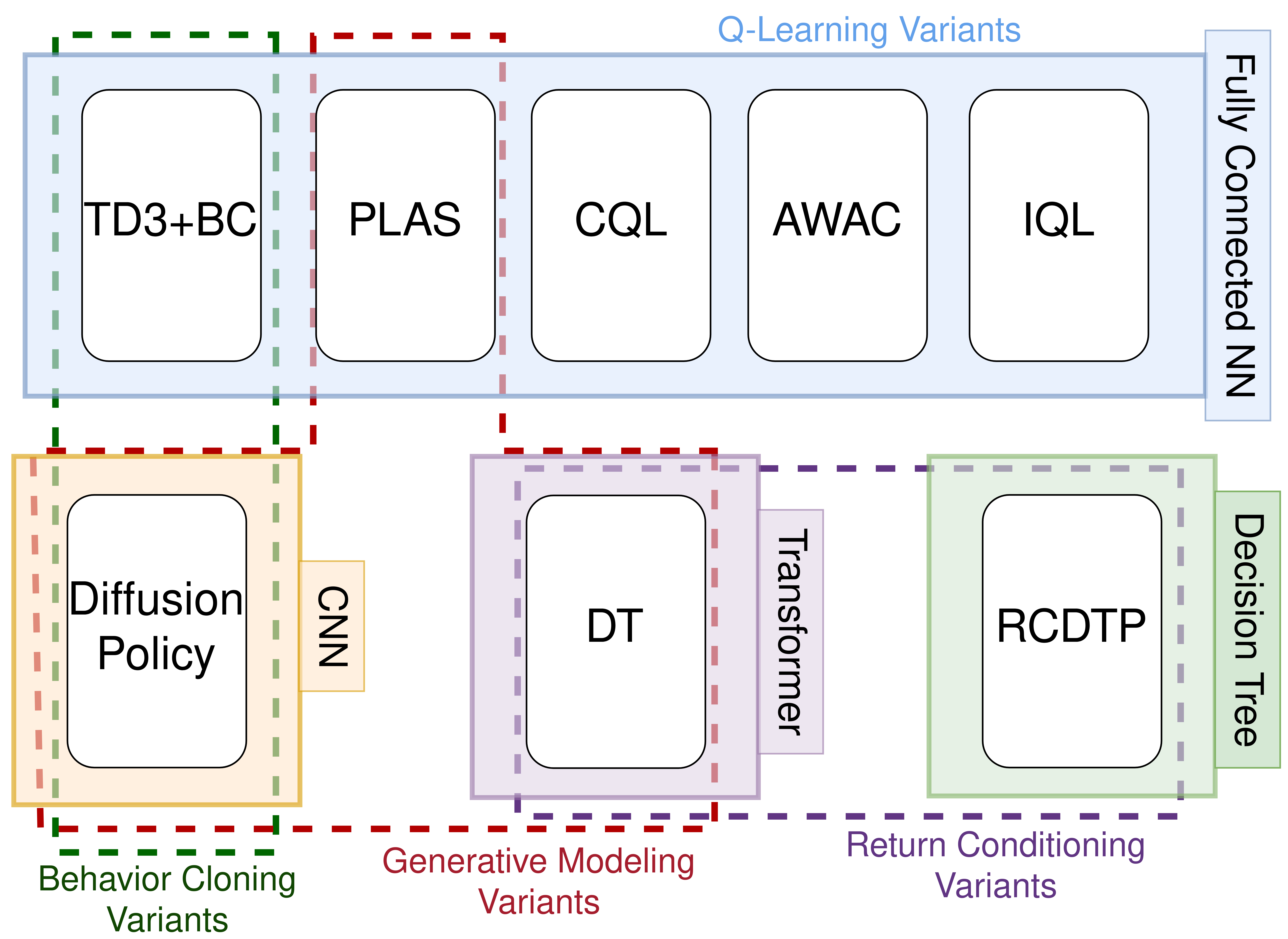}
    \caption{Overview of the Offline RL Methods used in this work}
    \label{fig:enter-label}
\end{figure}

\section{RELATED WORKS}
The authors of F1tenth evaluation virtual environment presented benchmark solutions employing static and online kinodynamic planning methodologies as means for continuous control. However, these solutions require a preexisting map for trajectory planning \cite{o2020f1tenthrl}.
The other popularly used controller designed using the Pure Pursuit algorithm guides a vehicle along a predefined racing line by continuously adjusting its steering towards a specific lookahead point on the trajectory \cite{conlter1992cmu}. By iteratively recalculating steering commands based on the selected point, it enables efficient and responsive path tracking. In our study, a method similar to pure pursuit is developed to collect the offline data that we use to train the offline RL algorithms.

Model Predictive Controllers tailored for small-scale racing cars offer effectiveness by generating dynamically feasible paths and optimizing performance in racing scenarios. However, these methods explicitly depend on both the map of the racing track and the intricate dynamics of the vehicle and also often require a predefined reference path for operation \cite{liniger2015optimization, kabzan2019learning, carrau2016efficient}. These meticulous prerequisites serve as a major distinction between the MPC-based controllers and their model-free counterparts in reinforcement learning.

Racing Dreamer uses a model-based deep reinforcement learning algorithm which requires learning the system dynamics (world model) using subcomponents that model the state representation, observation, reward and transition \cite{brunnbauer2022latent, hafner2019dream}. Further, their observation model for occupancy reconstruction method uses transposed convolutions to construct a 2D grayscale image and it is trained by providing patches of the true map that is required beforehand. Despite better sample efficiency and performance over model-free online algorithms, Racing Dreamer acknowledges significantly slower training, taking 6-8$\times$ longer than model-free methods \cite{brunnbauer2021model}.

In addressing the racing problem with Deep Q-Networks (DQN), as attempted in \cite{bosello2022traininaustria}, two key deviations (and shortcomings) can be noticed in their comparison with \cite{brunnbauer2022latent}. First, the requirement for discretization of the continuous action space into discrete categories such as forward, left, and right clearly presents a suboptimal approach, particularly in the context of real-world autonomous driving scenarios. Second, in contrast to the control inputs employed by \cite{brunnbauer2022latent}, which involve motor thrust and steering rate, \cite{bosello2022traininaustria} employs speed and steering angle as control inputs. 
% Additionally, \cite{bosello2022traininaustria} asserts the superiority of its algorithm over continuous action space algorithms used in \cite{brunnbauer2022latent} despite the deviation in the choice of control inputs.

Offline reinforcement learning, as a learning-based control framework, has proven effective across various domains such as robotics, finance, HVAC control, and recommendation systems \cite{tarasov2024corl, prudencio2023survey}. However, the application of offline RL within the context of racing is still unexplored, and our research seeks to fill this gap, aiming to chart pathways toward solving the autonomous racing problem with offline RL.

\section{METHODOLOGY}
\subsection{Data Collection \& Preparation}
\begin{figure} [h]
    \centering
    \includegraphics[width=0.9\linewidth]{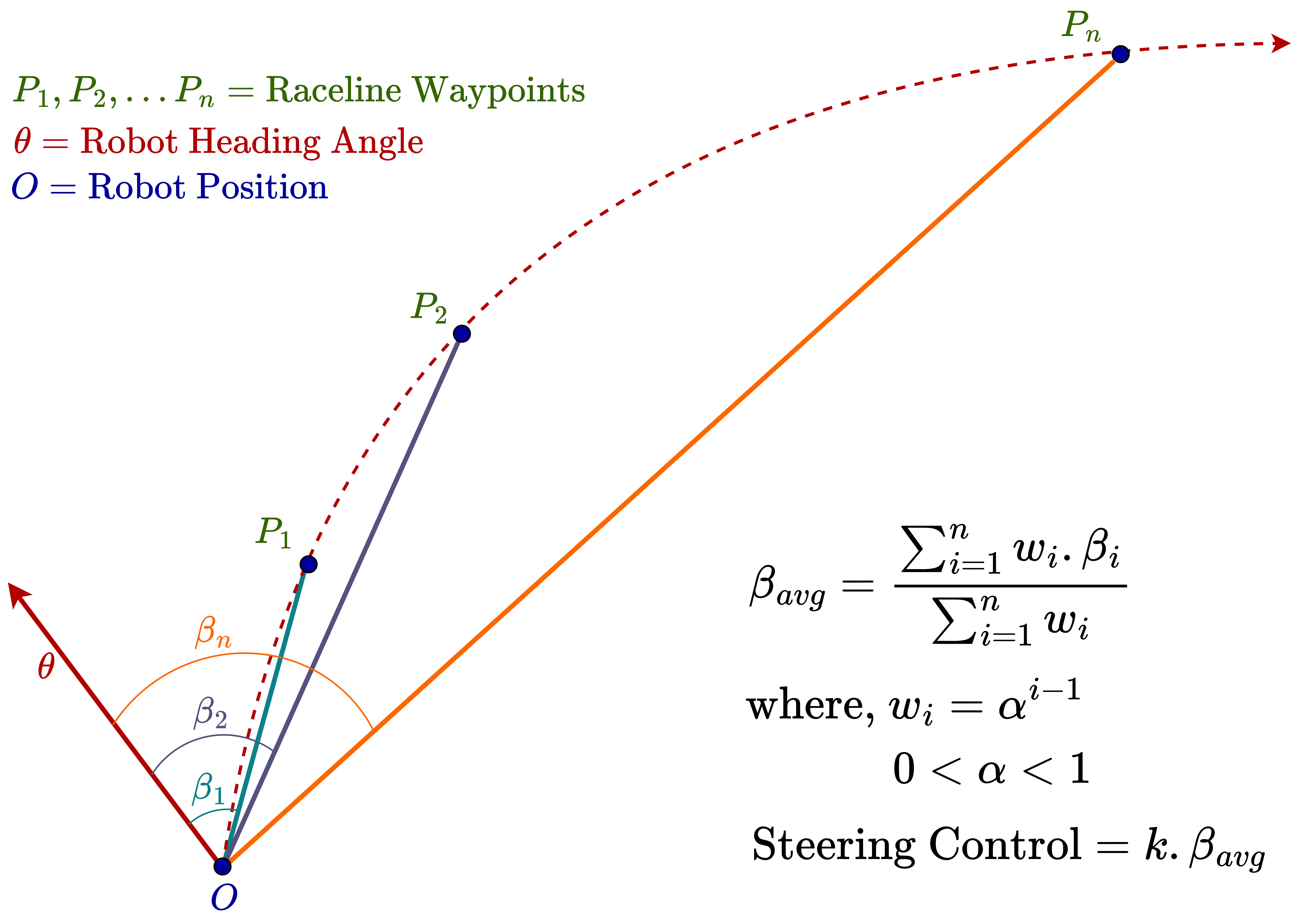}
    \caption{Waypoint based control of the f1tenth racecar}
    \label{Robot_Control_Waypoint}
\end{figure}

Offline data is collected by designing a controller that looks at a number of points ahead in the raceline. The error is the difference between the current heading angle and the angle made with the lookahead point. The final error angle is derived as a weighted mean of the errors with the lookahead points, wherein the contribution of distant points is mitigated by a designated discount factor. The steering command is proportional to this weighted mean error. Subsequently, the reference speed is computed as a function of the final steering error. %speed = max_speed * exp(k * steering_error). 
The motor command is proportional to the difference between this reference speed and the current speed of the vehicle. This approach, summarized in figure \ref{Robot_Control_Waypoint}, establishes a proportional controller mechanism to maintain the desired heading angle by steering adjustments and regulating speed through acceleration or deceleration.

\begin{figure} [ht!]
    \centering
    \includegraphics[width=\linewidth]{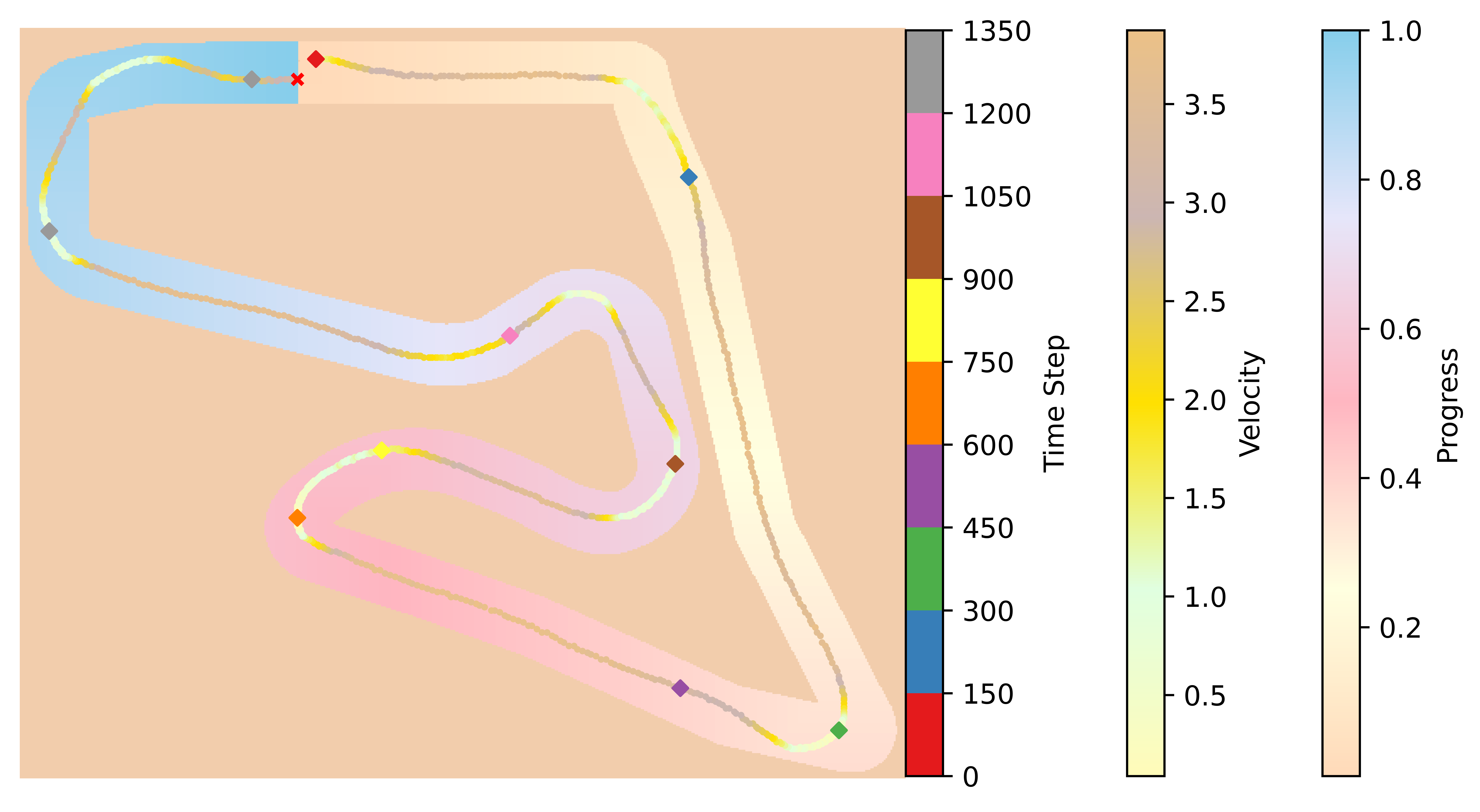}
    \caption{Visualization of an episode of f1tenth simulation in Austria racetrack}
    \label{austria_race_line velocity and progress}
\end{figure}

The racecar's simulation uses the aforementioned controller, enabling the collection of comprehensive data comprising observations, actions, reward feedback, and instances of termination/truncation across 100 episodes. Figure \ref{austria_race_line velocity and progress} illustrates an example of the racecar trajectory (one complete episode) simulated using this method. This concludes the offline dataset compilation essential for the subsequent training of our model. 

To train the return conditioned policies (eg. DT and RCDTP, which we will explain below in subsection $C$), we further preprocess the information at each timestep to determine the return-to-go value, which, together with the observation, constitute the input to these models (figure \ref{AgentDescription}). The return-to-go for a timestep is defined as the sum of future rewards from that timestep. 
\begin{align} \label{rtg equation 1}
    R_t = \sum_{t}^{T} {\gamma.r_t},  \quad \gamma=1 
\end{align}

\subsection{Environment} \label{Methodology Environment}

\subsubsection{Observation Space} \label{Environment Observation Space}
The observation space originally encompassed lidar data (1080 points spanning -135 degrees to +135 degrees), alongside pose, velocity and acceleration. The observation space for the RCDTP agent (\textit{and FCNN-based agents}) is condensed to 20 lidar beam distances evenly spaced between the -135 degree to +135 degree range and the vehicle’s speed information. Additionally, the present observation is augmented by stacking the past two observations. This helps the agent to infer contextual and positional information like heading direction, yaw rate, acceleration, etc and is prevalent in recent works on lidar-based racing \cite{bosello2022traininaustria, evans2023high}. 

The observation space for the agents employing the decision transformer and diffusion policy is, however, slightly different. Decision Transformer uses a context length ($K$, \textit{default 20} ) which, similar to sequence modeling, helps the transformer model during training to better discern the context in which the original policy could have generated the action in the dataset \cite{chen2021decision}. Diffusion Policy uses a similar hyperparameter $T_o$ (observation horizon, \textit{default 2}), denoting the latest steps of observations that are used to predict $T_p$ (prediction horizon) steps of immediate future actions \cite{chi2023diffusion}. It is therefore redundant in these models to further augment each observation with the past two observations, given that the models already have access to $K$ and $T_o$ steps of past lidar data and vehicle's speed. In our experiments, $K$ is set to 20, and $T_o$ is set to 3.

\subsubsection{Action Space}
The action space encompasses continuous control inputs, motor thrust and steering rate, within the range [-1, 1]. 

\subsubsection{Rewards}
Agent rewards are based on progress made along the racetrack. The agent incurs a penalty ($c=2$) and the episode terminates in the event of a collision; otherwise, it earns a reward equivalent to the progress made ($p_t$) \cite{brunnbauer2021model}. An agent that completes the lap without collision gathers a total reward of $\approx 100\%$, equivalent to the progress made $p_t=1$.
\begin{align}
r(t) &= \begin{cases} 
            -c & \text{if in collision} \\
            |p_{t} - p_{t-1}| & \text{otherwise} \\
\end{cases}
\end{align}

\begin{figure}[ht!]
\centering
\setkeys{Gin}{width=\linewidth}
    \begin{subfigure}[b]{0.45\linewidth}
        \centering
        \includegraphics{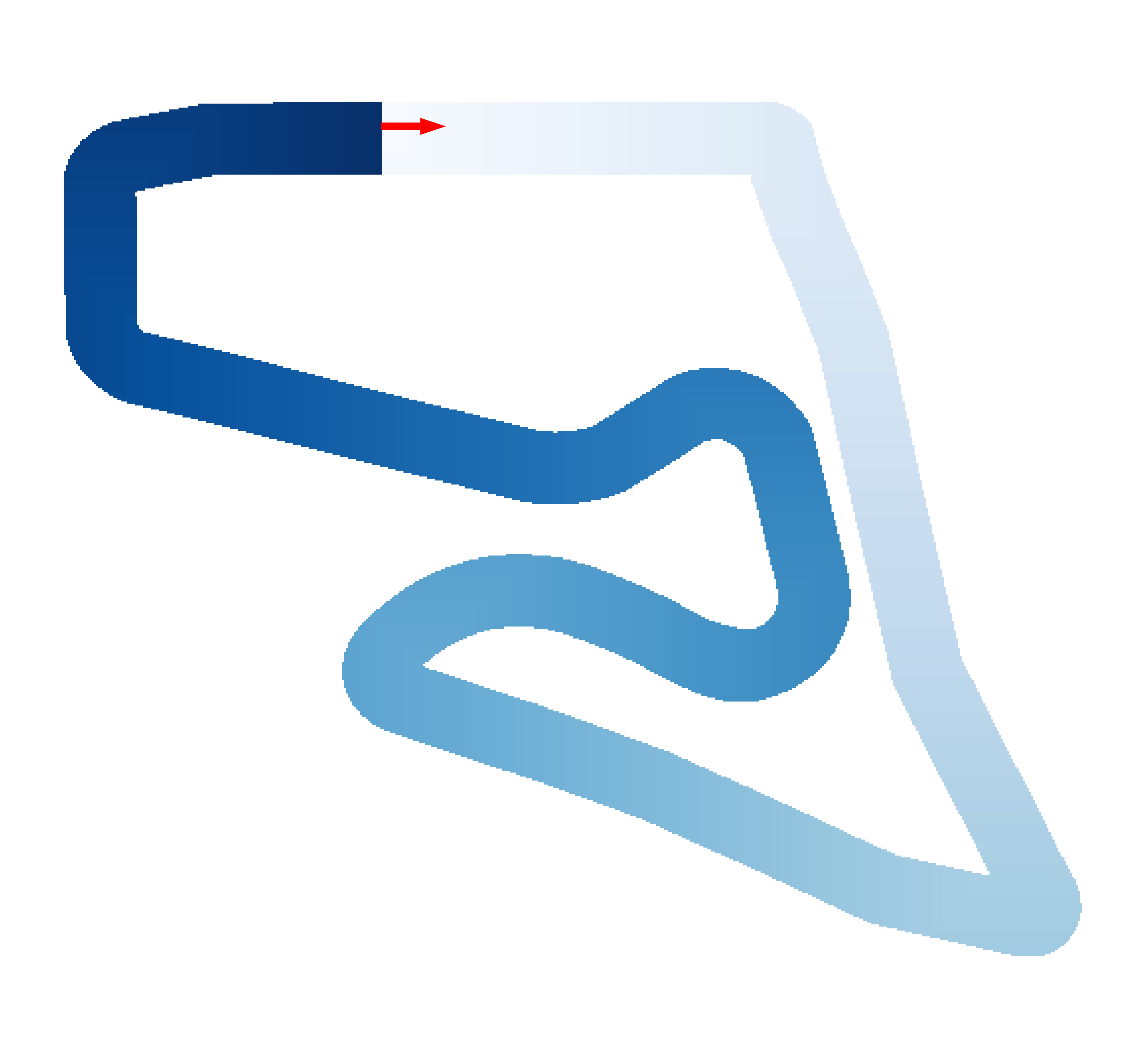}%{A.png}
        \caption{Austria}
        \label{fig:Austria Racetrack}
    \end{subfigure}
    \hfill
    \begin{subfigure}[b]{0.32\linewidth}
        \centering
        \includegraphics{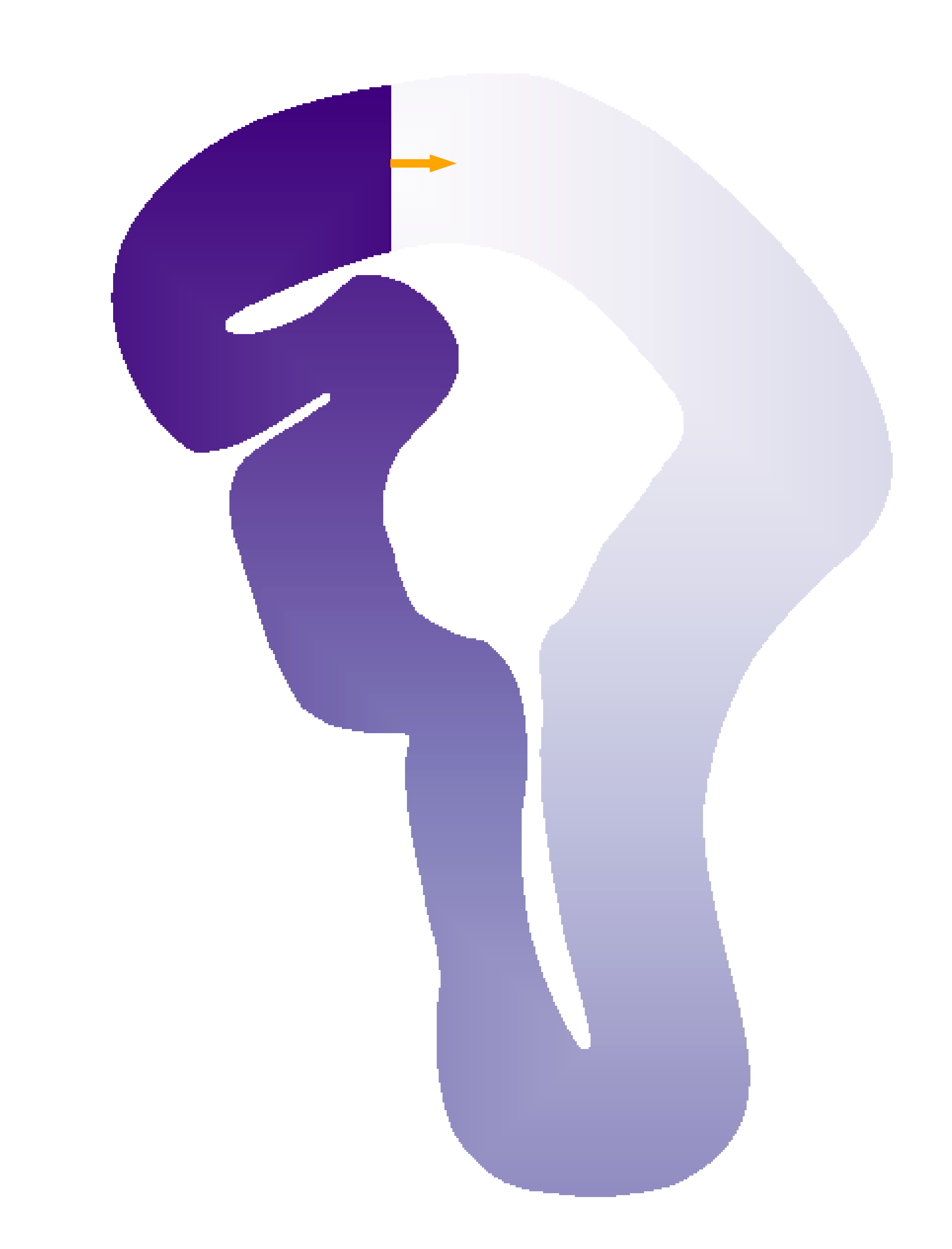}%{B.png}
        \caption{Berlin}
        \label{fig:Berlin Racetrack}
    \end{subfigure}
    \hfill
    \begin{subfigure}[b]{0.19\linewidth}
        \centering
        \includegraphics{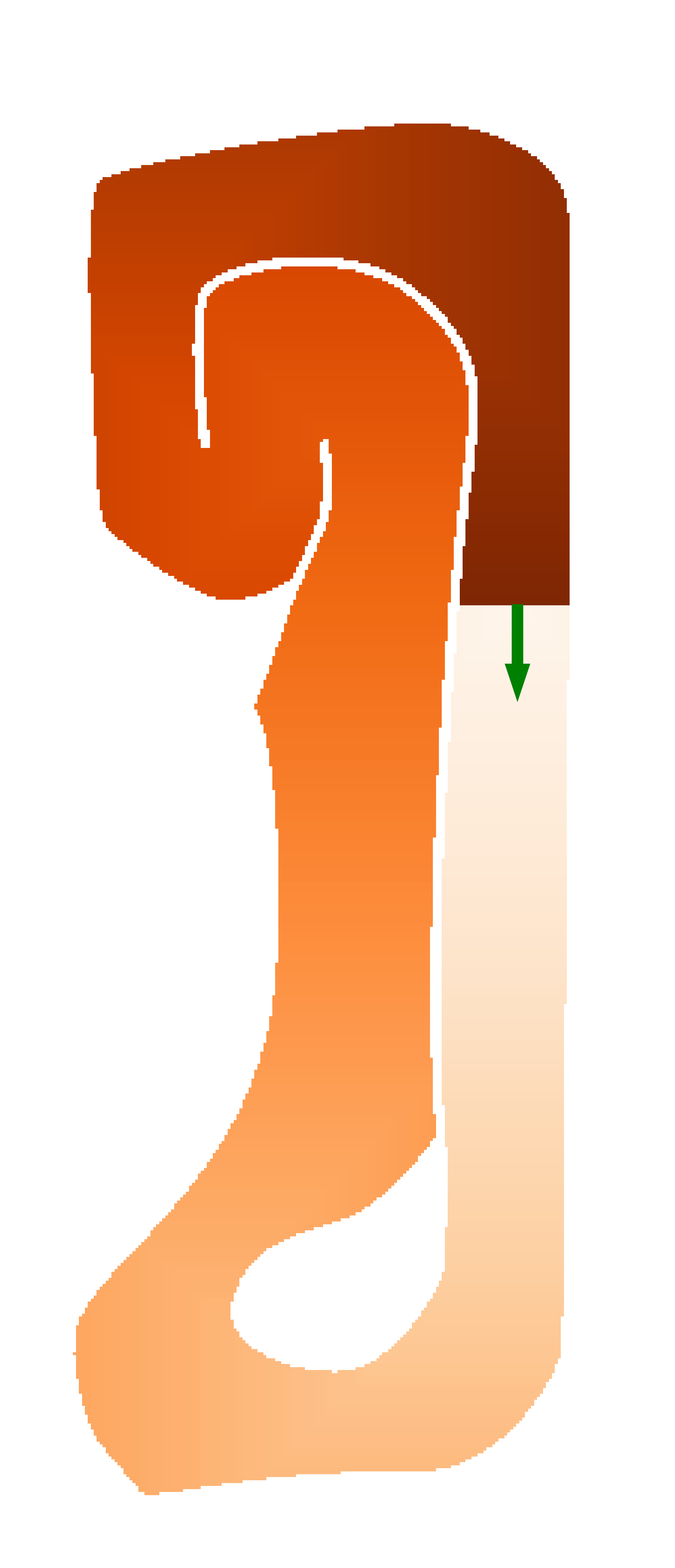}%{C.png}
        \caption{Torino}
        \label{fig:Torino Racetrack}
    \end{subfigure}
    \caption{Sample reward maps that are based on progress (Derived from the work of \cite{brunnbauer2021model, Axel2021racecar})}
\end{figure}

\subsection{Agent and Policy}
\subsubsection{Return Conditioned Decision Tree Policy (RCDTP)}
\begin{figure} [h]
    \centering
    \includegraphics[width=\linewidth]{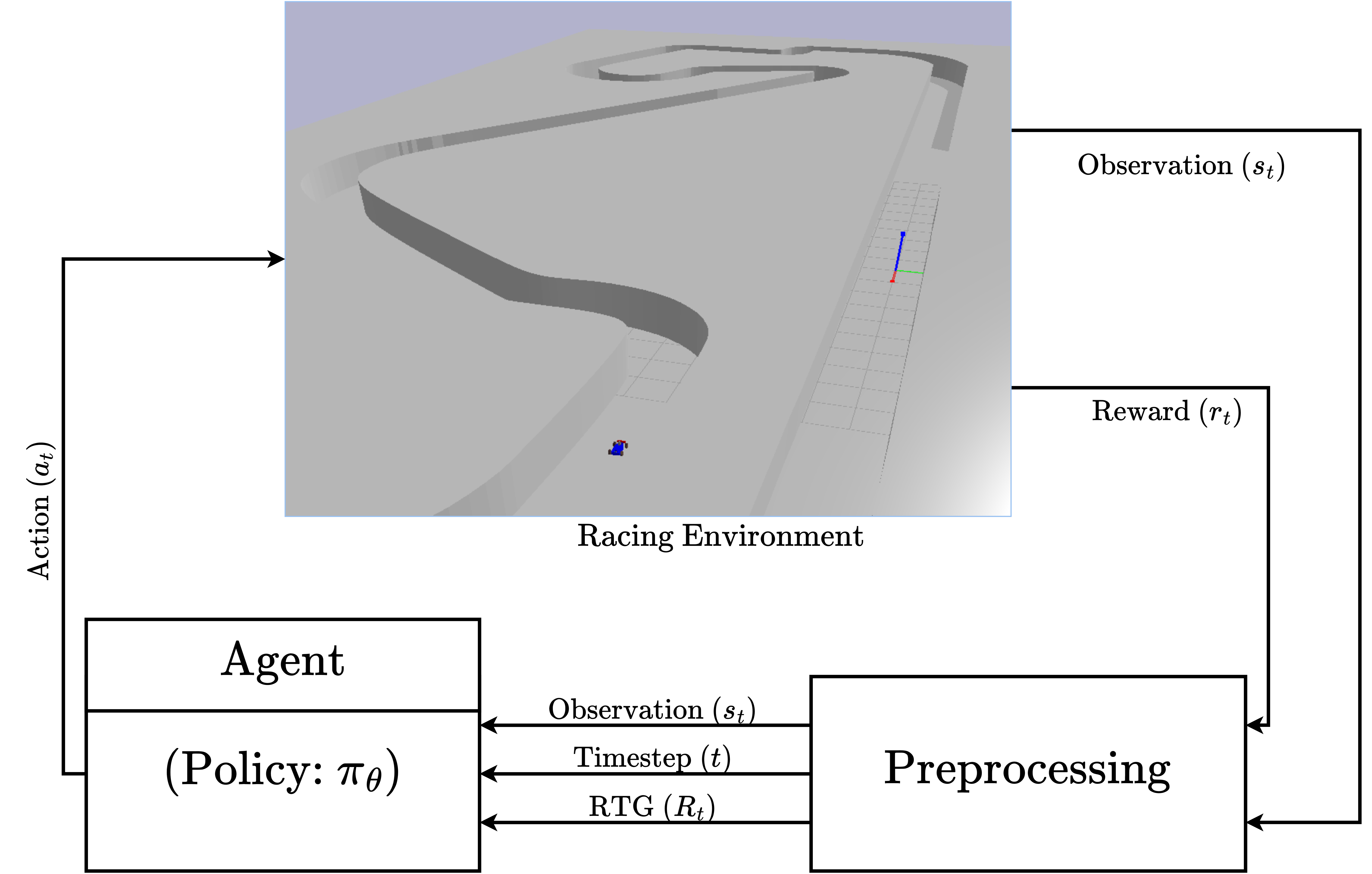}
    \caption{Agent-Environment interaction in a return conditioned setting}
    \label{AgentDescription}
    \vspace{-11pt}
\end{figure}

The agent's policy is deterministic, where the action is determined by three inputs: the observation, the current timestep, and the return-to-go (RTG) (figure \ref{AgentDescription}). 
$$
    \hat{a}_t = \pi(s_t, {R_t}, t; \theta)
$$
The policy-training is formulated as a regression problem. The objective seeks to minimize the sum of squared differences between the actual action ($a_n$) in the dataset and the actions predicted by the policy ($\hat{a}_n$). A tuple $<s_n, a_n, {R_n}, t_n>$ represents $n^{th}$ datapoint in the offline dataset of $N$ length.
\begin{align}
    J(\pi) &= \sum_{n=1}^N (a_n - \pi(s_n, {R_n}, t_n; \theta))^2      \label{meansquarederrorLoss}   \\ 
    \pi_\theta^*(s,{R_t}, t) &= \text{argmin}_{\pi_\theta} \sum_{n=1}^N (a_n - \pi(s_n, {R_n}, t_n; \theta))^2 \label{decisionTreepolicy}
\end{align}

A neural network aimed at minimizing the cost $J$ as described in equation \ref{decisionTreepolicy} typically employs gradient descent or a similar method, updating the model parameters $\theta$ through the iterative rule  $\theta \leftarrow \theta - \alpha \nabla_{\theta}J$. However, when using decision trees, the gradients and Hessians need to be calculated with respect to the model output ($\hat{a}$) at every boosting round \cite{chen2016xgboost}, rather than the model parameter $\theta$. 
The output of the decision tree associated with the $k^\text{th}$ boosting round, with observations at step $i$, is given as $\hat{a}_i^k:=\pi^k(s_i,R_i,i)$.
The final policy will consist of a sum of such outputs,
\begin{align}
    \pi(s,R,t)=\sum_{k=1}^K \pi^k(s,R,t),
\end{align}
for some fixed training budget, $K$, where each policy $\pi^k$ is referred to as a weak learner and is optimized according to the second-order approximation in equation \ref{second order approx xgb} and using standard techniques for decision-tree splitting and optimization \cite{xgboost-docs}.
\begin{align}
    \nabla_{\hat{a}^k_i} J &= \nabla_{\hat{a}^k_i} (a_i - \pi^{k}(s_i, {R_i}, t_i))^2 \\
    &= -2.(a_i - \pi^{k}(s_i, {R_i}, t_i)) \\
    \nabla_{\hat{a}^k_i}^2 J &= \nabla_{\hat{a}^k_i}^2   (a_i - \pi^k(s_i, {R_i}, t_i))^2 = 2
\end{align}

%\begin{align}
%    \nabla_{\hat{a}_i} J &= \nabla_{\hat{a_i}}\left( \sum_{k=1,2,..i,..N} (a_k - \pi(s_k, {R_k}, t_k))^2\right) \\
%    &= -2.(a_i - \pi(s_i, {R_i}, t_i)) \\
%    \nabla_{\hat{a}_i}^2 J &= \nabla_{\hat{a_i}}^2 \left(\sum_{k=1,2,..i,..N}  (a_k - \pi(s_k, {R_k}, t_k))^2 \right)= 2
%\end{align}

These gradients and Hessians are then utilized in constructing additional trees until convergence. 

\begin{align}
    \pi^k =  \text{argmin}\sum_{i=1}^N & \left[  \nabla_{\hat{a}^{k-1}}J \cdot \pi(s_i, {R_i}, t_i) + \dots \right.\nonumber\\ 
    &\left. \dots +\frac{1}{2}\nabla_{\hat{a}^{k-1}}^2 J \cdot \pi(s_i, {R_i}, t_i)^2\right]. \label{second order approx xgb}
\end{align}

\subsubsection{Decision Transformer (DT)}
In contrast to decision tree-based policies, which are trained through a straightforward regression problem and rely on a single vectorized information for each timestep, decision transformer policies undergo training as a sequence modeling problem. These policies use autoregressive modeling with GPT2 \cite{radford2019language} to derive actions based on historical trajectory sequence of past $K$ states, actions, returns, and timesteps (represented as positional embeddings) \cite{chen2021decision}.
\begin{align}
    \hat{a}_\tau = \pi( <s_t, a_t, {R_t}, t>_{t=\tau-K}^{t=\tau}; \theta) \label{decisionTransformerpolicy}
\end{align}
where $<\cdot>_l^u$ indicates a sequence of tuples.

In Decision Transformer training with a given dataset of offline trajectories, minibatches are sampled with a sequence length equal to the context length ($K$). The prediction head associated with the input tuple ($<s_t, a_t, R_t, t>$) is trained to predict $\hat{a_t}$ using mean-squared error loss as in equation \ref{meansquarederrorLoss}.

\subsubsection{Diffusion Policy (DP)}
Except for Diffusion Policy, all other models used in this work predict action for the immediate timestep and the action is executed at that particular timestep. Diffusion Policy, in contrast, borrows ideas from receding horizon control to realize closed loop action-sequence prediction with temporally consistent long-horizon planning. The model predicts $T_p$ (prediction horizon) steps of immediate future actions, of which $T_a$ (action horizon, $T_a \leq T_p$) steps are executed before the next inference/prediction. 
\begin{align}
    <\hat{a}>_{t=\tau}^{t=\tau+T_p} = \pi( <s_t>_{t=\tau-T_o}^{t=\tau}; \theta) \label{decisionTransformerpolicy}
\end{align}
For a total of M denoising iterations with DDPMs, the action sequence $<\hat{a}>_{t=\tau}^{t=\tau+T_p}$ is obtained by iteratively denoising a Gaussian noise sample $<a^M_t>$ into $<a^0_t>$. This involves subtracting the output of the noise prediction network $\varepsilon_\theta(<s_t>, <a^m_t>,m)$ M times iteratively, where $m$ represents the $m^{th}$ denoising iteration \cite{chi2023diffusion}. 
% The training loss is the modified form of standard DDPM MSE loss of the predicted noise.

\subsubsection{Q-learning Variants}
In offline reinforcement learning (RL) for continuous-space control tasks, the Q-learning variants typically adopt an actor-critic style, wherein the actor learns the policy and the critic learns the value function \cite{konda1999actor}. However, the distributional shift, caused by the disparity between the data distribution and the learned policy distribution, poses a significant challenge in using `online’ methods in the offline settings  directly \cite{levine2020offline}. To address this, the offline RL algorithms often employ various strategies such as using regularization methods to constrain the policy and/or the critic, or sometimes altogether avoiding training the model outside the support of the distribution of the dataset.

\begin{itemize}

    \item Conservative Q Learning (CQL) builds upon SAC \cite{haarnoja2018softactor} and aims to induce pessimism by minimizing the \textit{value} for out-of-distribution actions while trying maximizing it for in-distribution ones.
    
    \item Advantage-Weighted Actor-Critic (AWAC) uses an advantage-weighted regression objective, similar to AWR \cite{peng2019advantage}, for policy improvement. AWAC, however, makes use of off-policy critic (Q-function) to estimate the advantage, and this enhances its sample efficiency. 
    
    \item In PLAS, the policy is learned in the latent action space using variational autoencoders, inherently constraining the policy to select actions from within the distribution. We augment PLAS with an additional perturbation layer to generalize out-of-distribution actions.
    
    \item TD3+BC extends the online algorithm TD3 \cite{fujimoto2018addressing} with an additional behavior cloning (BC) regularization, encouraging the policy to predict in-distribution actions.

    \item Implicit Q-learning (IQL) incorporates training an additional state-value network so as to avoid querying out-of-distribution actions during the training of the Q-network. The policy is then trained using the advantage-weighted regression method.
    
\end{itemize}

% Conservative Q Learning (CQL) builds upon SAC \cite{haarnoja2018softactor} and aims to induce pessimism by minimizing the \textit{value} for out-of-distribution actions while trying maximizing it for in-distribution ones.

% Advantage-Weighted Actor-Critic (AWAC) uses an advantage-weighted regression objective, similar to AWR \cite{peng2019advantage}, for policy improvement. AWAC, however, makes use of off-policy critic (Q-function) to estimate the advanatge, and this enhances its sample efficiency. 

% In PLAS, the policy is learned in the latent action space using variational autoencoders, inherently constraining the policy to select actions from within the distribution. We augment PLAS with an additional perturbation layer to generalize out-of-distribution actions.

% TD3+BC extends the online algorithm TD3 \cite{fujimoto2018addressing} with an additional behavior cloning (BC) regularization, effectively encouraging the policy to predict in-distribution actions.

% Implicit Q-learning (IQL) incorporates training an additional state-value network so as to avoid querying out-of-distribution actions during the training of the Q-network. The policy is then trained using the advantage-weighted regression method.

\section{RESULTS}
Following the data collection and offline training procedures detailed in the preceding section, this section visually and quantitatively compares the performance of our approach (RCDTP) against more recent models such as Decision Transformer and Diffusion Policy, as well as other fully connected neural network (FCNN) based offline models. 

\subsection{SOTA Online RL Techniques as ``non''baselines}
\begin{figure} [ht!]
    \centering
    \includegraphics[width=\linewidth]{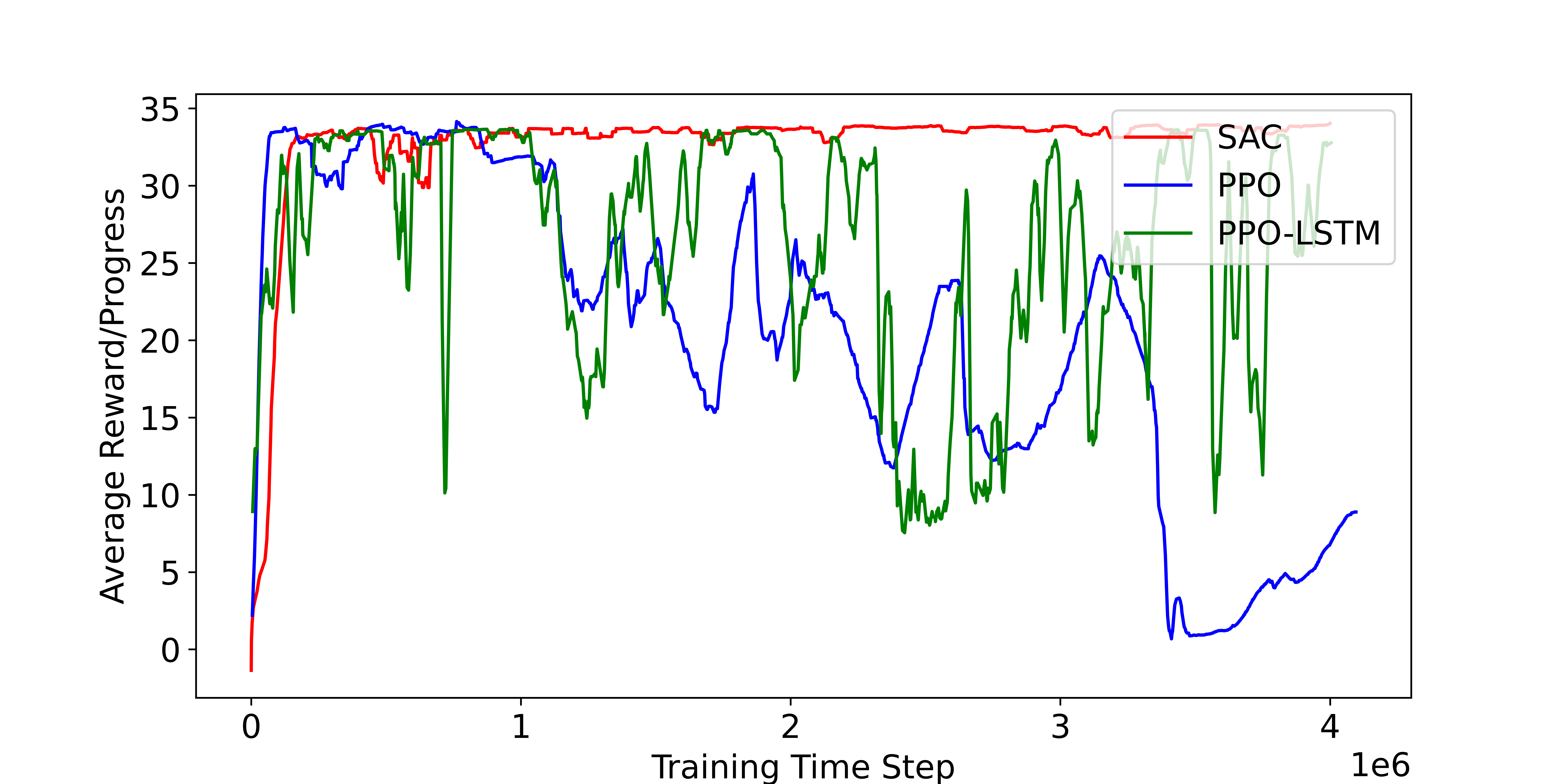}
    \caption{Online training log in Austria racetrack (Stable Baselines3 \cite{stable-baselines3} implementations % of these algorithms 
    were used.)}
    \label{online_training_austria}
    \vspace{-11pt}
\end{figure}

At the outset of this research, one desideratum was to compare these offline, regression-based learning techniques to state-of-the-art (SOTA) online reinforcement learning algorithms. However, attempts to train an agent {\em online} in the Austria environment proved unsuccessful, as evident in the training log (figure \ref{online_training_austria}) where all the agents fail to surpass the $35\%$ progress mark. These state-of-the-art online algorithms used the same simulation environment (including observation and action spaces) as described in \ref{Methodology Environment} for RCDTP.

\begin{figure*}[ht!]
\centering
\setkeys{Gin}{width=\textwidth}
    \begin{subfigure}[b]{0.99\linewidth}
        \centering
        \includegraphics{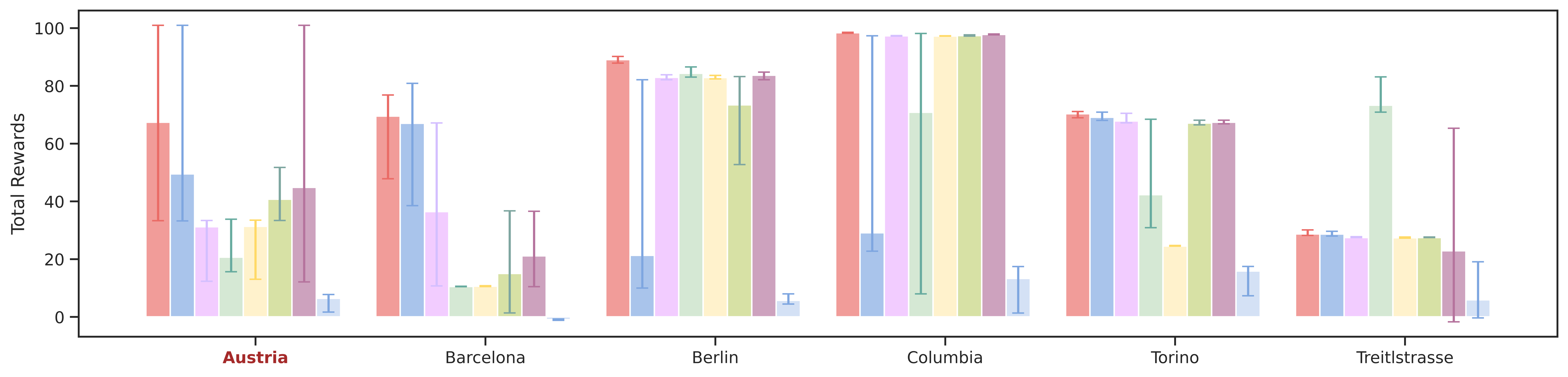}
        \vspace{-9mm}
        \caption{}
        \label{TrainedInAustria_XGB_vs_DT}
    \end{subfigure}
    \hfill
    \begin{subfigure}[b]{0.99\linewidth}
        \centering
        \includegraphics{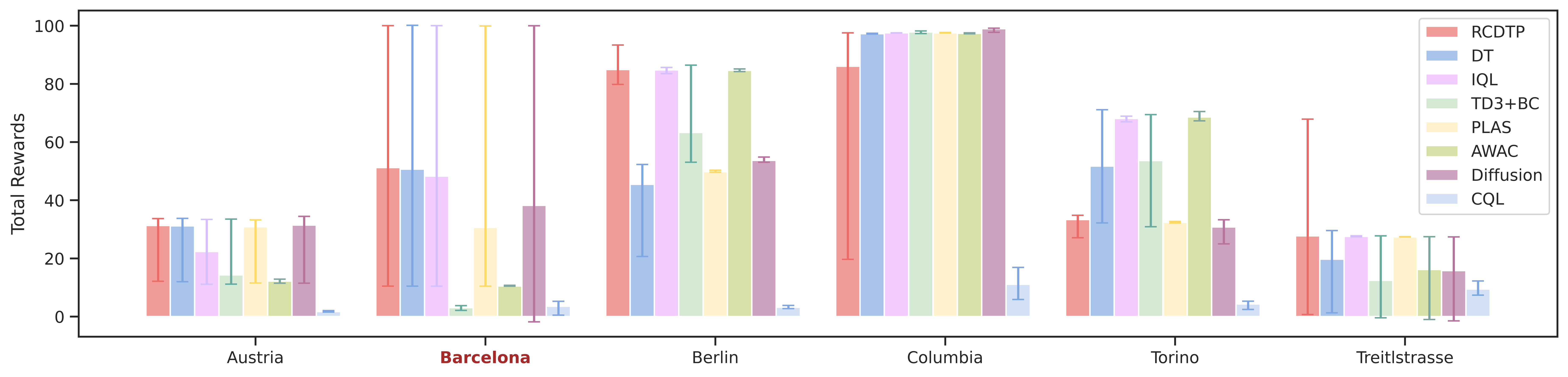}
        \vspace{-9mm}
        \caption{}
        \label{TrainedInBarcelona_XGB_vs_DT}
    \end{subfigure}
    \begin{subfigure}[b]{0.99\linewidth}
        \centering
        \includegraphics{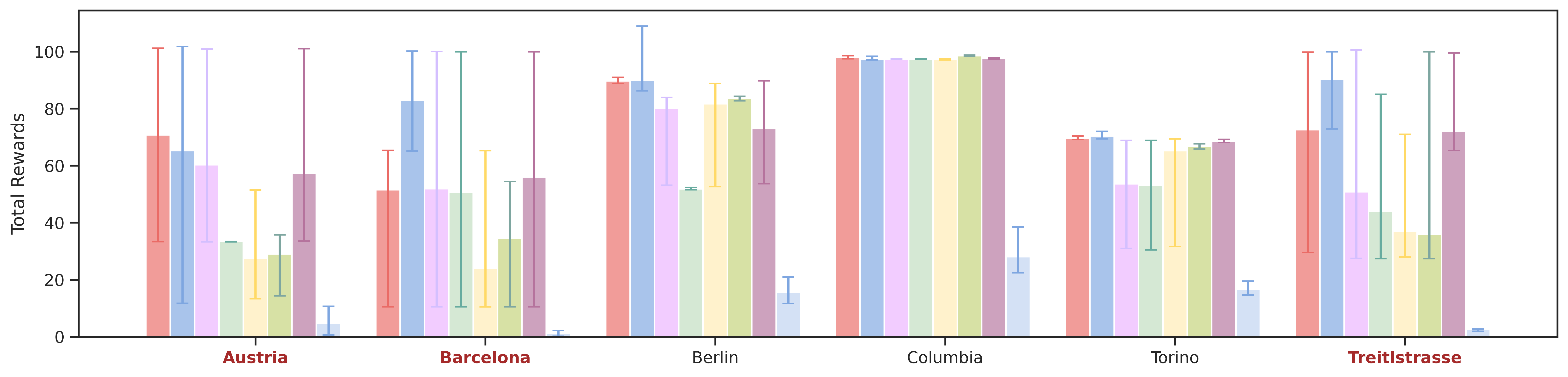}
        \vspace{-9mm}
        \caption{}
        \label{TrainedInMixedData_XGB_vs_DT}
    \end{subfigure}
    \vspace{-2mm}
    \caption{Total episode rewards when trained in (a). Austria (b). Barcelona (c). Austria-Barcelona-Treitlstrasse 
    (The highlighted racetrack names in each figure represent the `home' environment used to collect the training dataset.)}
    \vspace{-8pt}
\end{figure*}

\subsection{Generalization performance with 1-Track training}
Figure \ref{TrainedInAustria_XGB_vs_DT} shows the results when the training is conducted in the Austria racetrack dataset, and subsequent evaluations are performed in different racing environments, including Austria itself. The bar chart illustrates the total return of an episode averaged over 10 episodes of testing, and the whisker extensions denote the maximum and minimum returns observed across these episodes. The performance of all the agents trained using different algorithms are juxtaposed with RCDTP performance for comparison.

Despite being exposed to only 100 episodes of expert demonstration only on the Austria racetrack, the offline models--especially RCDTP--demonstrate competent performance on other unseen racetracks as well. 
This seen2unseen transfer of the agents' proficiency can also be attributed to the inherent difficulty of the Austria racetrack, as it includes sharper left and right turn scenarios. Nevertheless, the models' performance on the Austria racetrack itself is throttled by a challenging right turn when the racecar attains approximately $33-35\%$ of the maximum progress. We saw in our experiments that this can be remedied for RCDTP by sufficiently increasing the number of stacked past observations (by default set to 2 as described in section \ref{Environment Observation Space}). Although this approach (stack of 50 obs \& 3000 boosting rounds) increases the average total returns to as high as $90\%$ in the designated track (Austria), there are two major drawbacks: the cross-track generalization ability of the RCDTP agent suffers (potentially overfitting) and its quick training characteristic capability can be compromised due to increased dimensionality of input features.

Interestingly, the performance of many RL agents trained solely on the Barcelona racetrack dataset exhibited a decrease when assessed in various racetracks, including the `home' environment Barcelona itself (figure \ref{TrainedInBarcelona_XGB_vs_DT}). Although most of these agents were able to sometimes accumulate ($\approx100\%$) in the Barcelona racetrack, the average returns for most agents is less than when they were trained in only-Austria dataset.
Similarly, when they were assessed in the Austria track, all of them failed to cross the $\approx35\%$ mark. The performance of the agents in racetracks like Berlin and Torino also mostly declined.

\subsection{Generalization performance with Multi-Track training} 
When all models are trained on a dataset comprising 100 `expert' demonstrations in each of the racetracks -- Austria, Barcelona, and Treitlstrasse -- the decision transformer-based agent shows significant improvement, often surpassing the RCDTP agent in terms of lap completion rate and total returns. This superior performance of DT over other methods can also be partly attributed to the advantageous access to the past $K$ observations facilitated by the extended context length. The average episodic return accumulated by each agent in the six tracks can be visualized in figure \ref{TrainedInMixedData_XGB_vs_DT}. Note that the training datasets for experiments corresponding to figures \ref{TrainedInAustria_XGB_vs_DT} and \ref{TrainedInBarcelona_XGB_vs_DT} are subsets of that corresponding to \ref{TrainedInMixedData_XGB_vs_DT} (Au-Ba-Tr). The incorporation of datasets from other racetracks actually helped Decision Transformer and Diffusion policy agents to improve performance in Austria and Barcelona compared to when exposed to the dataset from Austria or Barcelona alone. However, the average score accumulated by the RCDTP agent in these racetracks remains largely unchanged with this exposure to additional datasets, implying a lack of generalized learning from other racetrack datasets. 

While other Q-learning variants utilizing traditional feed-forward neural networks demonstrated decent performance with default hyperparameters and minimal tuning, CQL posed considerable challenges in achieving effectiveness even with tuning efforts.  Such hypersensitivity to hyperparameter choices and fragility regarding implementation details have been noted as an important observation in \cite{tarasov2024corl}. In our experiments, CQL training was characterized by blow-ups in the conservative loss and the critic loss.

\subsection{Robustness Analysis}

\begin{figure} [ht!]
    \centering
    \includegraphics[width=\linewidth]{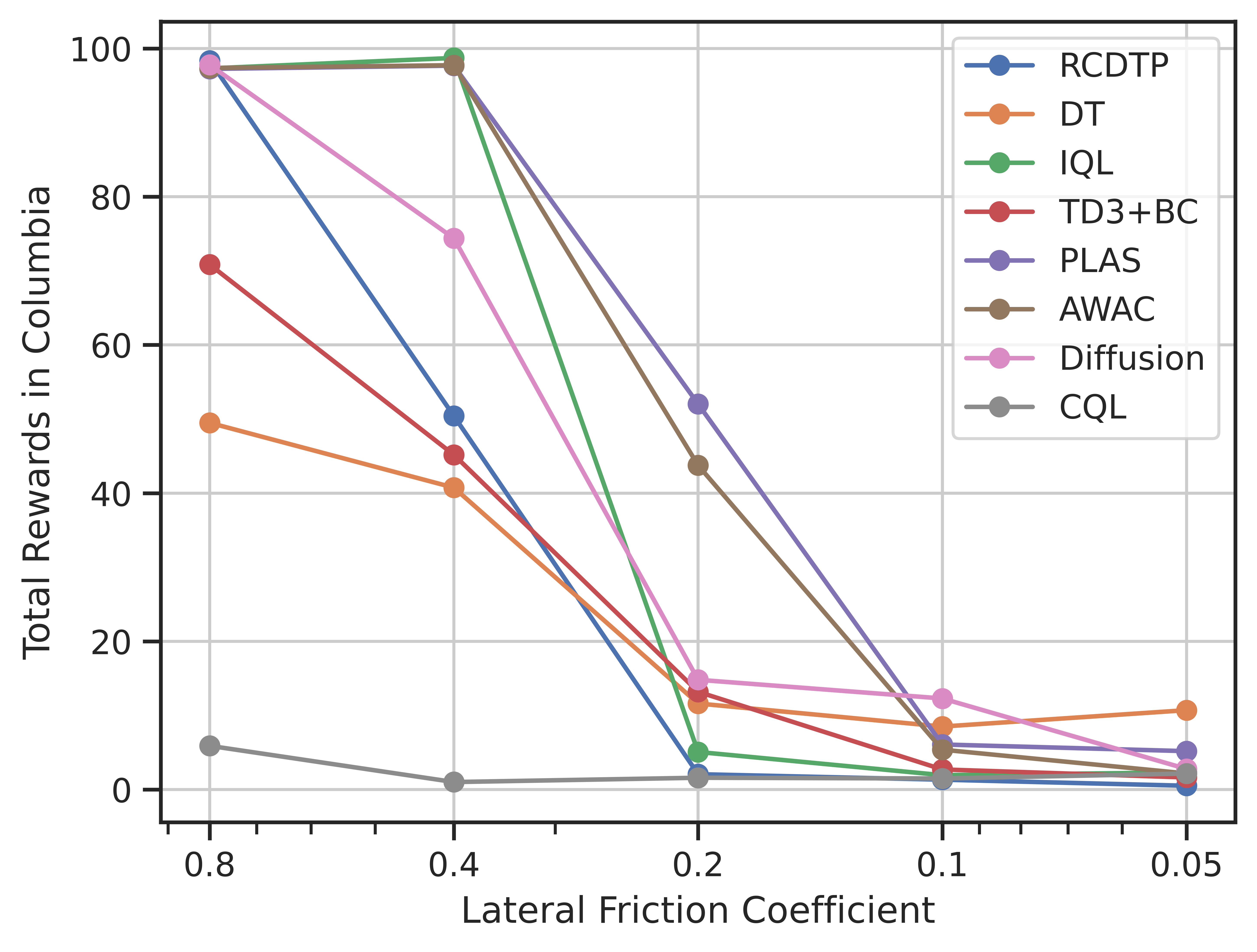}
    \caption{Performance variation of agents trained in Austria racetrack dataset across different lateral friction coefficients in the unseen Columbia racetrack}
    \label{FrictionCoefficientInColumbia}
    \vspace{-11pt}
\end{figure}

%As an experiment 
To assess the adaptability and robustness, the agents trained exclusively in the Austria racetrack were deployed and tested in the Columbia racetrack under varying lateral friction coefficients. This experiment was designed to simulate changing road conditions encountered in autonomous driving scenarios, where factors such as weather conditions, rainfall, or snowfall may alter road friction. The coefficient was systematically decreased from its default value of 0.8, with the agents' cumulative rewards recorded and averaged over 10 episodes on the Columbia racetrack (figure \ref{FrictionCoefficientInColumbia}). Despite the RCDTP agent generally outperforming most of other agents across all racetracks (as evident in figure \ref{TrainedInAustria_XGB_vs_DT}), including Columbia, it exhibited the most significant performance drop with changing environment dynamics.  Interestingly, methods based on generative models (PLAS, DT, DP) emerged as the most resilient, exhibiting the least performance drop and maintaining superior performance until the lowest friction coefficient value. Among the FCNN-based methods, PLAS and AWAC exhibited similar performance, with PLAS slightly better, over the decreases in friction coefficient. %AWAC maintained the best performance besides PLAS over the decreased friction coefficient.

\subsection{Training Time Comparision}
Finally, besides differences in racetrack performance, the models also exhibit trade-offs between model complexity and computational efficiency, guiding the selection of an appropriate model for offline reinforcement learning in F1tenth racing environments (Table \ref{tab:Training and inference performance across different racetrack datasets}). While an RCDTP model can be trained within a few minutes on a CPU (Intel Core i9-13900KF), the larger models take significantly more time. Decision Transformer takes around 30 to 35 times more training time on a powerful GPU (NVIDIA RTX 4090 - 24 GB) and Diffusion Policy may take up to 200 times more training time on the same GPU device. 

Our experiments show that both Decision Transformer and Diffusion Policy need increased training time as the dataset size grows. For RCDTP, an increase in dataset size leads to a requirement for more boosting rounds, consequently prolonging the training time. However, for Q-learning-based offline RL methods, the training time remains relatively consistent when the same epoch length is used. Among these methods, PLAS boasts the shortest per-epoch training time, while AWAC requires the most.

% incomplete
\begin{table} [hbt]
    \centering
    \caption{Comparison of training time across different racetrack datasets}
    \label{tab:Training and inference performance across different racetrack datasets}
    \begin{tabularx}{\linewidth}{|p{1.4cm}|p{0.85cm}|p{1.3cm}|p{0.9cm}|p{0.7cm}|X|}
        \hline
         Trained In & Model & Rounds/ Epochs & \multicolumn{3}{c|}{Training Time ($\mu$, $\sigma$, Dev)  }\\
         \hline
         \hline
         Austria & RCDTP  & 10000 & 57.14s & 3.31s & CPU  \\
         \hline
         & DT & 100000 & 1987.55s & 12.54s & GPU  \\
         \hline
         & IQL & 60 & 2394.3s & 56.4s & GPU  \\
         \hline
         & TD3+BC & 50 & 1699.3s & 50.9s & GPU  \\
         \hline
         & PLAS & 70 & 1267s & 85.5s & GPU  \\
         \hline
         & AWAC & 50 & 4098s & 56.5s & GPU  \\
         \hline
         & DP & 400 & 13640s & 241.64s & GPU  \\
         \hline
         & CQL & 20 & 2247.7s & 68.6s & GPU  \\
         \hline
         \hline
         Barcelona & RCDTP & 15000 & 112.20s & 4.29s & CPU\\
         \hline
         & DT & 100000 & 3347.67s & 162.3s & GPU \\
         \hline
         & IQL & 60 & 2355.3s & 71.4s & GPU  \\
         \hline
         & TD3+BC & 50 & 1687.7s & 31.2s & GPU  \\
         \hline
         & PLAS & 70 & 1304s & 42.1s & GPU  \\
         \hline
         & AWAC & 50 & 4156s & 72.3s & GPU  \\
         \hline
         & DP & 400 & 18935s & 374.6s & GPU  \\
         \hline
         & CQL & 20 & 2285.3s & 52.6s & GPU  \\
         \hline
         \hline
         Aus-Bar-Trt & RCDTP & 20000 & 363.29s & 12.32s & CPU \\
         \hline
         & DT & 100000 & 9425.81s & 102.11s & GPU \\
         \hline
         & IQL & 60 & 2421.3s & 71.2s & GPU  \\
         \hline
         & TD3+BC & 50 & 1721.3s & 51.21s & GPU  \\
         \hline
         & PLAS & 70 & 1292.3s & 53.5s & GPU  \\
         \hline
         & AWAC & 50 & 4122s & 76.3s & GPU  \\
         \hline
         & DP & 400 & 38869s & 165.4s & GPU  \\
         \hline
         & CQL & 20 & 2310.3s & 72.5s & GPU  \\
         \hline
    \end{tabularx}
    \vspace{-11pt}
\end{table}

\section{CONCLUSIONS}
In conclusion, this study demonstrated the efficacy of different offline reinforcement learning algorithms to `learn’ expert policies in the F1tenth racing environment. The observed zero-shot transferability from seen to unseen environments suggests a valuable baseline and method for tackling new racetracks with a working suboptimal policy trained in a different track.
Conversely, during the training of {\em online} RL algorithms for the same progress maximization task in a racetrack, a common trend emerged: each online algorithm plateaued at a suboptimal return value, failing to complete the lap. Among the algorithms that were assessed, SAC showed the most stable {\em online} training.

Among the offline RL algorithms, the Return Conditioned Decision Tree Policy (RCDTP) based agent exhibited notable sample efficiency in learning when the dataset consisted of data only from the same track. It mostly outperformed others in both seen and unseen environments, while also demonstrating a significantly faster training (within minutes in CPU). However, when it came to a larger and more varied dataset involving diverse racetracks, the decision transformer exhibited its ability to learn a more general policy, benefiting from the sequence modeling capabilities inherent in the transformers. Among the FCNN-based variants, PLAS not only showed the smallest per epoch training time but also adapted best to the change in lateral friction coefficient in an unseen racetrack environment.

Furthermore, the findings underscore the distinct use cases of these algorithms. RCDTP, which uses XGBoost, proves ideal for sample-efficient training and quick experimentation with a consistent environment or racetrack dataset. On the other hand, larger models like Decision Transformer and Diffusion Policy excel in scenarios with larger, cross-track datasets, and adapt better to dynamics change, showcasing their ability to generalize and learn robust policies. Diffusion Policy, nonetheless, has two main drawbacks: it demands extensive training time and compute resources, and it relies on an expert-only dataset. These insights suggest a strategic approach to deploying the offline algorithms tailored to the specific problem at hand. For instance, in a situation suiting return-conditioned models, a rapid deployment of Decision Trees-based policy can efficiently collect data across various tracks, and these datasets can then be leveraged to train larger models like Decision Transformer for a more generalized and potentially superior policy applicable to a broader spectrum of racetracks and racetrack conditions. This concept can be extended beyond F1tenth racing, offering implications for learning-based control tasks across diverse domains.

Future research topics include but are not limited to multi-agent scenarios, sim2real applications and safe reinforcement learning. 
The research of multi-agent scenarios, where agents learn to competitively race against each other as well as navigate the race track, can offer valuable insights into cooperative and competitive dynamics of the agents. 
Similarly, sim2real research offers potential to bridge the gap between simulated and real-world racing while enhancing the applicability and robustness of learning-based methods in practical settings. The exploration of safe RL methods is also equally crucial to ensure the safety of autonomous racing agents in real-world scenarios. Engineering different reward functions that encourage faster lap completion while enforcing safety constraints can contribute to the deployment of more reliable autonomous racing agents.

%%%%%%%%%%%%%%%%%%%%%%%%%%%%%%%%%%%%%%%%%%%%%%%%%%%%%%%%%%%%%%%%%%%%%%%%%%%%%%%%

\bibliographystyle{IEEEtran}
\bibliography{IEEEabrv,bib}

% \begin{thebibliography}{99}
% \bibitem{c20} J. P. Wilkinson, ÒNonlinear resonant circuit devices (Patent style),Ó U.S. Patent 3 624 12, July 16, 1990. 
% \end{thebibliography}

\end{document}